\documentclass[10pt,letterpaper]{article}
\usepackage[top=0.85in,left=2.75in,footskip=0.75in]{geometry}

\usepackage{amsmath,amssymb}

\usepackage{changepage}

\usepackage[utf8x]{inputenc}

\usepackage{textcomp,marvosym}

\usepackage{cite}

\usepackage{nameref,hyperref}

\usepackage{graphicx}
\usepackage{booktabs}       
\usepackage{url}
\usepackage{tikz}

\usepackage[final]{changes}

\usepackage[right]{lineno}

\usepackage{microtype}
\DisableLigatures[f]{encoding = *, family = * }

\usepackage{array}

\usepackage{multirow} 
\newcolumntype{L}[1]{>{\raggedright\arraybackslash}p{#1}}

\newcolumntype{C}[1]{>{\centering\arraybackslash}p{#1}}

\newcolumntype{R}[1]{>{\raggedleft\arraybackslash}p{#1}}

\newcolumntype{+}{!{\vrule width 2pt}}

\newlength\savedwidth

\def\@fnsymbol#1{\ensuremath{\ifcase#1\or *\or \dagger\or \ddagger\or
   \mathsection\or \mathparagraph\or \|\or **\or \dagger\dagger
   \or \ddagger\ddagger \else\@ctrerr\fi}}
\newcommand{\ssymbol}[1]{$^{\@fnsymbol{#1}}$}

\def\transform#1{\url{#1}\hskip 0pt plus 1pt}

\def\hyphenateAndTtWholeString #1{\xHyphenate#1$\wholeString\unskip}

\def\xHyphenate#1#2\wholeString {\if#1$%
    \else\transform{#1}%
    \takeTheRest#2\ofTheString\fi}

\def\takeTheRest#1\ofTheString\fi
{\fi \xHyphenate#1\wholeString}

\def\urlx #1{\href{#1}{\hyphenateAndTtWholeString{#1}}}

\usepackage{pifont}

\usepackage[aboveskip=1pt,labelfont=bf,labelsep=period,justification=raggedright,singlelinecheck=off]{caption}

\makeatletter
\renewcommand{\@biblabel}[1]{\quad#1.}
\makeatother

\usepackage{lastpage,fancyhdr,graphicx}
\usepackage{epstopdf}
\fancyheadoffset[L]{2.25in}
\fancyfootoffset[L]{2.25in}
\newcommand{\gcmark}{\textcolor{green}{\ding{51}}}%
\newcommand{\rxmark}{\textcolor{red}{\ding{55}}}%
\newcommand{\svr}{\cite{varga2019no1}}
\newcommand{\lstm}{\cite{varga2019no2}}
\newcommand{\deepbiq}{\cite{bianco2018use}}
\newcommand{\multi}{\cite{varga2020multi}}
\newcommand{\deeprn}{\cite{varga2018deeprn}}
\newcommand{\svrt}{CNN-SVR}
\newcommand{\lstmt}{CNN-LSTM}
\newcommand{\deepbiqt}{DeepBIQ}
\newcommand{\multit}{MultiGAP-NRIQA}
\newcommand{\deeprnt}{DeepRN}

\begin{document}
\vspace*{0.2in}

\begin{flushleft}
{\Large
\textbf\newline{Critical analysis on the reproducibility of visual quality assessment using deep features} 
}
\newline
\\
Franz G\"otz-Hahn\textsuperscript{1*},
Vlad Hosu\textsuperscript{1\ddag},
Dietmar Saupe\textsuperscript{1\ddag}
\\
\bigskip
\textbf{1} Dept. of Computer and Information Science, Universit\"at Konstanz, Konstanz, Baden-W\"urttemberg, Germany
\\
\bigskip

%
%

\ddag These authors contributed equally to this work.




* franz.hahn@uni.kn

\end{flushleft}
\section*{Abstract}
Data used to train supervised machine learning models are commonly split into independent training, validation, and test sets. \replaced{This paper illustrates}{In this paper we illustrate} \replaced{that complex data leakage cases}{that intricate cases of data leakage} have occurred in the no-reference image and video quality assessment literature. Recently, papers in several journals reported performance results well above the best in the field. However, our analysis shows that information from the test set was inappropriately used in the training process in different ways and that the claimed \added{performance} results cannot be \replaced{achieved}{reached}. When correcting for the data leakage, the performances of the approaches drop even below the state-of-the-art by a large margin. Additionally, we investigate end-to-end variations to the discussed approaches, which do not improve upon the original.

\section*{Introduction}
\label{sec:intro}
The goal of visual quality assessment is to provide a quantitative measure of multimedia perceptual quality, particularly for digital images and video sequences. Many applications in multimedia strive to optimize the quality of their outputs. For example, in image and video coding, the best reconstruction of an original input image or video sequence after the encoding and decoding process is desirable under a given bitrate constraint. In most cases, the consumers of multimedia products are people, and thus, the perceptual media quality is the most important type of visual quality~\cite{brunnstrom2013qualinet}. Therefore, in the last couple of decades, many image and video quality datasets have been introduced. The subjective quality annotations contained therein have been gathered in elaborate laboratory experiments and, more recently, by crowdsourcing. Subsequently, much effort has been invested in engineering so-called objective quality measures that automate the extraction of visual quality from digital multimedia objects, i.e., without collecting subjective ratings from human observers. For this purpose, the benchmark datasets of subjective ratings provide the means for designing the mathematical models that predict the ratings. Furthermore, these models are intended to generalize to other multimedia data not contained in the training sets.

Modern algorithms that predict an image or video's visual quality depend on machine learning techniques such as support vector regression or deep neural networks, trained in a supervised fashion. For this purpose, benchmark datasets are commonly split three-fold into a training, validation, and test set. The training set is understood as ``a set of examples used to fit the parameters of a classifier'' \cite{ripley2007pattern} or regressor. The validation set is used as a stopping criterion of the learning process and a means for selecting the optimal model hyperparameters, such as the number of layers or hidden units in a neural network layer. A model with its parameters fitted on the training set is tested on the validation set to estimate its generalization performance. While the validation set is chosen to produce an unbiased initial estimation of the generalization performance, with every evaluation and change in the hyperparameters towards the optimal performance, more information about the data in the validation set is incorporated into the model.

Finally, the test set performance remains the only unbiased estimate. It is used solely for the performance evaluation of a fully trained model. This means, only once the model's configuration is finalized, the test set is employed to determine the model's generalization performance and, therefore, its real-world applicability. For this purpose, it is paramount that adjustments of a machine learning model are never based on its evaluation on the test set, as the test set's purpose is then lost by influencing the model's performance on itself.

To ensure this procedure's validity, data is commonly first sampled randomly without replacement according to some ratio for the test set, followed by the validation set in the same fashion. The remaining data is left for training. For small datasets, a typical split is 60/20/20\% for training, validation, and test, respectively, while larger datasets often use smaller validation and test set sizes. A quality predictor's performance is then primarily measured by the Pearson linear correlation coefficient (PLCC) or the Spearman rank-order correlation coefficient (SROCC) of the predictions with the ground-truth qualities in the corresponding benchmark datasets.

Reproducibility and explainability are machine learning topics that gained increased traction in recent years. Various surveys have shown that a vast majority of papers do not make their code available ~\cite{hutson2018artificial,gundersen2018state}. Nearly half of them do not include pseudocode either. Moreover, the simple inclusion of pseudocode does not guarantee reproducibility~\cite{raff2019step}. 

Despite the efforts mentioned above to validate and test independently, data leakage is considered by many experts as one of the biggest problems in machine learning. It is a primary culprit for irreproducibility. Data leakage in machine learning relates to training a model on information that should only be available at test time. One of the simplest ways data leakage can occur is when the target itself is used as an input to the model. However, data leakage can manifest in machine learning in many subtle ways, such as being introduced in several different training procedure stages. \added{In fact, one could argue that the definition of data leakage could be elaborated to mean any kind of information flow between data used in any part of the machine learning pipeline, such as information of the validation set being available at training time. This kind of data leakage should conceptually not improve the performance on an independent test set, but in nature it is a problem very similar to what is understood as classical data leakage.} \replaced{Therefore, it becomes increasingly difficult}{This is incredibly difficult} to spot data leakage when multiple processing steps are involved, or statistical information is extracted during pre-processing. 

For example, if one considers the entirety of a dataset when normalizing it, the information would be leaked between the training, validation, and test sets. Afterward, performing cross-validation might unintentionally change the estimated performance on the validation and test sets, although the impact might be relatively small. Instead, one \replaced{should}{would} first normalize the training set, represent the transformation parameters independent of the data, and apply the same parameterized transformation to the validation and test sets before other learning algorithms are used.

Generally, state-of-the-art machine learning approaches for image and video quality assessment (IQA and VQA) are marked by small, incremental improvement. In contrast, five recent papers showed remarkable progress for deep learning models for IQA and VQA and certainly deserve special attention in the field. In this contribution, we provide a study on the validation and reproducibility of these existing findings. However, our results turn out to be negative in that the existing findings are found to be irreproducible. The problems with the questionable contributions stem from adequately training machine learning models to predict data and validating their expected performance correctly.

Our findings deserve public attention. Firstly, such false claims of considerable advances of the state-of-the-art will discourage researchers from pursuing those small incremental steps vitally important to experimental research. Secondly, papers submitted for publication that yield such incremental improvements, however below the presumed but false state-of-the-art, are likely to be rejected.

We share and discuss these five data leakage cases in the visual quality assessment domain in this communication. We summarize the papers that proposed similar image and video quality prediction approaches and describe the subtle ways data leakage caused overly optimistic results that do not hold under scrutiny and careful reimplementation. Specifically, we discuss three image quality assessment (IQA) publications in \deepbiq{}, \deeprn{}, and \multi{}, as well as \svr{} and \lstm{} from the video quality assessment (VQA) domain. We henceforth refer to the approaches described in these publications as \deepbiqt{}, \deeprnt{}, \multit{}, \svrt{}, and \lstmt{}, respectively, in order to increase readability. 


\begin{center}
  \begin{tabular}{R{2cm}C{0.75cm}L{2.2cm}L{4cm}L{2.2cm}}
    Method & Ref. & Author & Paper Title & Venue \\
    \cmidrule[1pt]{1-5}
    \deepbiqt{} & \deepbiq{} &  Bianco S., Celona L., Napoletano P., Schettini R. & On the use of deep learning for blind image quality assessment & Signal, Image and Video Processing \\
    \rule{0pt}{3ex} \deeprnt{}  & \deeprn{}  &  Varga D., Saupe D., Szir\'anyi T. & DeepRN: A content preserving deep architecture for blind image quality assessment & International Conference on Multimedia and Expo \\
    \rule{0pt}{3ex} \multit{} & \multi{} &  Varga D. & Multi-pooled inception features for no-reference image quality assessment & Applied Sciences \\
    \rule{0pt}{3ex} \svrt{} & \svr{} & Varga D. & No-Reference Video Quality Assessment Based on the Temporal Pooling of Deep Features & Neural Processing Letters \\
    \rule{0pt}{3ex} \lstmt{} & \lstm{} &  Varga D., Szir\'anyi T. & No-reference video quality assessment via pretrained CNN and LSTM networks & Signal, Image and Video Processing \\
    \cmidrule[1pt]{1-5}
  \end{tabular}
\end{center}

We report the corrected results for the approaches, the reverse-engineering process, and reconstruct the mistakes that have likely resulted in the incorrect published performance numbers. The complete code necessary to reproduce the results in this report is available online.\footnote{
See \urlx{https://github.com/FranzHahn/NPL-50-3-2595-2608-Correction},  \urlx{https://github.com/FranzHahn/SIVP-13-8-1569-1576-Correction}, \urlx{https://github.com/FranzHahn/MDPI-10-6-2186-Correction}, and \urlx{https://github.com/FranzHahn/SIVP-12-2-355-362-Correction}. There we also included links to MATLAB workspace binaries containing trained networks, extracted features, as well as non-aggregated results.}
From our analysis, we conclude the following:
\begin{enumerate}
    \item The \deepbiqt{} method does not yield the level of performance claimed in the paper \deepbiq{}. On LIVE-in-the-wild, an IQA dataset, the SROCC on test sets are only $0.76 \pm 0.02$, instead of the 0.89 claimed, and on TID 2013, an artificially degraded IQA dataset, the model performs with $0.64 \pm 0.05$ SROCC, instead of the 0.96 claimed in the paper. The discrepancy between these results and the true values can be attributed to a case of data leakage, where the model's fine-tuning step illegitimately had access to data from the test set. 

    \item In both \svrt{} and \lstmt{}, which are very close adaptations of \deepbiqt{} for the field of VQA, a similar case of data leakage as in \deepbiqt{} occurs, despite contrasting claims. Furthermore, an additional case of data leakage occured, where the validation set used for fine-tuning was not properly separated from the training set. They, therefore, do not yield the performances as claimed. On KoNViD-1k, a large-scale VQA dataset, the SROCCs on test sets are only $0.67 \pm 0.04$ and $0.63 \pm 0.05$, respectively, instead of 0.85. 

    \item \multit{}, an enhanced version of \deepbiqt{} suffers from a different kind of data leakage causing illegitimate performance values for two artificially degraded IQA datasets, KADID-10k, and TID2013. On KADID-10k, the SROCCs on test sets are only $0.81 \pm 0.05$, instead of 0.97 as claimed.

    \item The published performance of \deeprnt{} also cannot be reproduced. The introduction of the simple types of data leakage the author revealed in personal communication to have happened does not explain the published results.

    \item Finally, we present alternative end-to-end solutions for \svrt{} and \multit{}. We show that na\"ively fine-tuning Inception-style networks is not a promising solution for the visual quality assessment domain in general.
\end{enumerate}

The rest of this paper is organized as follows. First, we summarize the broad approach that all papers in question have in common, outlining the major differences and representing their performance results. Next, we describe the different kinds of data leakages we discovered in a nearly chronological order, as later publications alleviated some problems of earlier works and introducing new types of data leakage. We begin with the discussion of data leakage Case I, occurring in the fine-tuning step of \deepbiq{}. Then, we report Case II appearing in the fine-tuning step of \svr{} and \lstm{}. Further, both \svr{} and \lstm{} are affected by data leakage Case III. The similar but subtly different type of data leakage Case IV occurring in \multi{} is covered last. Finally, we discuss the approaches more generally and with respect to our findings, before finishing the article with a performance analysis of fine-tuning Inception-style networks that are an alternative to the approaches suggested in \deepbiq{}, \multi{}, \svr{}, and \lstm{}.

\section*{Visual quality assessment based on deep features}
\label{sec:vqaondeepfeatures}

For the design of many image and video processing methods and their practical use, objective assessment of perceptual characteristics, such as quality or aesthetics, may be required. In order to develop such algorithms, benchmark datasets have been created that contain multimedia items together with annotated perceptual attributes. These labels usually are mean opinion scores (MOS) from lab-based or crowdsourced user studies. They serve as ground truth data for model evaluation, as well as for training and validation of machine learning approaches.

In recent years, deep convolutional neural networks (DCNN) have seen increased use as feature extraction tools. Specifically, internal activations of individual DCNN layers have been useful for human perception tasks in general, despite the original training criterion only considering object classification. For example, in aesthetics quality assessment, several works~\cite{hii2017multigap,hosu2019effective} have represented images as internal activations of Inception-style networks. Others have used similar concepts for image and video quality prediction~\cite{gao2018blind,gotz2019no}, or perceptual similarity assessment~\cite{zhang2018unreasonable}. 

\begin{figure}[t!]
    \centering
    \includegraphics[width=0.99\columnwidth]{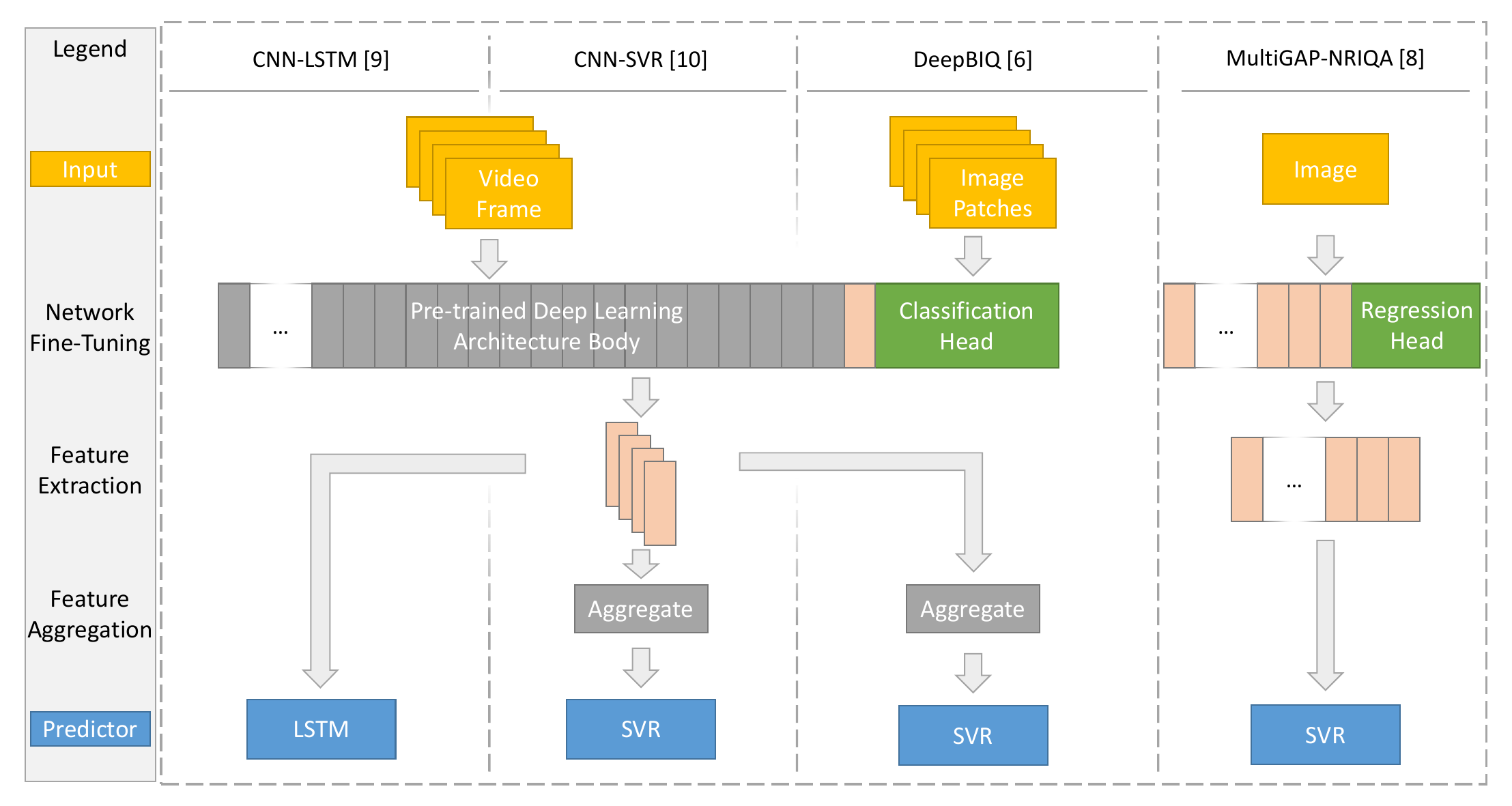}
    \caption{This is a high-level flowchart of the procedures employed in four of the five referenced visual quality papers. Video frames, image patches, or images are input into a pre-trained deep learning network with a classification or regression head replacement. The entire network is fined-tuned and then used as a feature extractor. The approaches differ by using the last layer or all layers as a feature source. The feature representations are then aggregated, where appropriate, and used to train the final quality predictor.}
    \label{fig:flowchart}
\end{figure}

Recently, this approach has been utilized for no-reference image and video quality assessment in \deepbiq{}, \svr{}, \lstm{}, \multi{}, and \deeprn{}, that are under investigation in this contribution. The concept of the first four very similar approaches can be broken down into the three stages of fine-tuning, feature extraction, and quality prediction. \added{Depending on the final predictor, an additional step of feature aggregation may be required}. Figure \ref{fig:flowchart} is a high-level representation containing the broader differences between the methods, \added{ranging from different inputs to the network, to the features that are extracted from the network, as well as the way features are aggregated to serve as an input to the final regressor}. In the following, we will first describe the three separate stages and outline the difference and then discuss the differences to \deeprn{} separately. For clarity, whenever we reference frames, we are relating it to \svrt{} and \lstmt{}, whereas the term images refers to \deepbiqt{}, \deeprnt{} and \multit{}.

\subsection*{Comparison of approaches}
\label{sec:vqa:method}

The general approach for \lstmt{}, \svrt{}, \deepbiqt{}, and \multit{} can be broken down to three stages. In the following we illustrate the differences at each stage, respectively.

The first stage common to all four methods is the fine-tuning of a pre-trained DCNN network. While the publications utilize and compare different feature extraction networks, the choice makes no difference from a conceptual standpoint. The last layers of the networks used for the original task of object classification are replaced with a new head to accommodate the task of visual quality assessment. In the case of \lstmt{}, \svrt{}, and \deepbiqt{} a 5-way softmax layer is added, which distinguishes between five visual quality classes derived from quantizing the MOS. Alternatively, in \multit{}, a regression head is used, omitting quality classes as a proxy and directly predicting the MOS score. The inputs to the fine-tuning process are resized and cropped frames from a video, a set of random image patches, or resized and cropped individual images.

After fine-tuning, video frames, image patches, or images are passed through the network and the activations of a selection of layers within the network are extracted as feature representations. Here, \multit{} considers all Inception modules of the Inception-style network and performs global average pooling to reduce the spatial dimensions. The other three approaches, however, only consider the last pooling layer as a feature representation. In the case of \lstmt{}, the sequential frame-level features are saved. In contrast, for \svrt{}, the frame-level features from the same video are aggregated by performing average, median, minimum, or maximum pooling to obtain video-level representations. In the case of \deepbiqt{} the patch-level features from the same source image are aggregated by average pooling to obtain image-level representations.

Finally, the extracted feature vectors serve as inputs to a regressor. The aggregated video-level features in \svrt{}, the aggregated image-level features in \deepbiqt{}, or the individual image features in \multit{} are used to train a support vector regressor (SVR). In contrast, \lstmt{} leverages the power of long-short term memory (LSTM) networks, which process sequences of data points. Therefore, all the frame-level features of an individual video are used as an input. This approach can potentially retain temporal cohesion of the changes of features during the playback of a video, improving the prediction performance over an aggregated approach, such as \svrt{}.

In contrast, \deeprnt{} has a slightly different albeit related approach to IQA. Similar to the previous methods the first step is comprised of fine-tuning a pre-trained DCNN network with a classification head. However, once the network is fine-tuned, the classification head is again replaced with a spatial pyramid pooling layer followed by a small fully connected network. Spatial pyramid pooling maps an input of arbitrary size to a fixed size, allowing the use of fully connected layers, on the one hand, and the evaluation of images of arbitrary resolution on the other. The output of the model used in \deeprnt{}, finally, is a distribution of ratings, rather than absolute MOS scores, as in the previous approaches. In theory this could leverage additional information about the input.

\subsection*{Implementation details and performance}
\label{sec:vqa:implementation}
The authors evaluate their approaches on well-established video and image quality datasets. According to \svrt{} and \lstmt{}, the evaluation is performed on KoNViD-1k~\cite{hosu2017konstanz}. In the case of \deepbiqt{} both authentic and artificial IQA datasets are considered with LIVE-in-the-wild~\cite{ghadiyaram2016massive} and TID2013~\cite{ponomarenko2015image}, respectively. \multit{} considers two more modern artificially distorted IQA datasets KADID-10k~\cite{kadid10k} and TID2013~\cite{ponomarenko2015image}, as well as the modern in-the-wild IQA dataset KonIQ-10k~\cite{hosu2020koniq}. For \svrt{}, \lstmt{}, and \multit{} the best performance was achieved using an Inception-V3 network architecture as a baseline feature extraction network. Alternatively, \deepbiqt{} considers the dated CaffeNet~\cite{jia2014caffe}, a slightly modified AlexNet~\cite{krizhevsky2012imagenet} variant, as a feature extractor. As a performance metric, the correlation \added{coefficient} between the model predictions and the ground truth MOS is reported, \added{which is} a common metric for I/VQA algorithms.

For \svrt{}, the peak average performance on test sets from KoNViD-1k was given by a Pearson linear correlation coefficient (PLCC) of 0.853 and a  Spearman rank-order correlation coefficient (SROCC) of 0.849. In the case of \lstmt{} the final performance reported was 0.867 PLCC and 0.849 SROCC. In both accompanying papers \svr{} and \lstm{}, another dataset (LIVE-VQA) was also used; however, we focus on the former in our discussion here for brevity and simplicity. The previous best-reported performance on KoNViD-1k had been achieved by TLVQM~\cite{korhonen2019two}, with a 0.77 PLCC and 0.78 SROCC, respectively. The improvement in the performance of 0.08/0.10 PLCC and 0.07/0.07 SROCC is substantial, considering the field's usually incremental improvements.

In the IQA domain, \deepbiqt{} was evaluated on different pre-train setups and patch-level feature aggregation methods. Ultimately, final performance was achieved using pre-trained weights from a hybrid of ImageNet~\cite{krizhevsky2012imagenet} and Places~\cite{zhou2014learning}, as well as prediction pooling, meaning that the predictions on all patches derived from the same image are averaged to obtain the image-level prediction. In the case of LIVE-in-the-wild, the previously best reported performance was given by FRIQUEE~\cite{ghadiyaram2014blind,ghadiyaram2016massive} with 0.71/0.68 PLCC/SROCC, while \deepbiqt{} claimed to improve this to 0.91/0.89 PLCC/SROCC. Additionally, the approach is compared to related works on a variety of artificial datasets. For brevity we will only consider their results on TID2013, the most modern in the comparison. Here, \deepbiqt{} claimed state-of-the-art performance of 0.96 for both PLCC and SROCC. In their own evaluation they cite HOSA~\cite{xu2016blind} as the next best competitor with 0.96/0.95 PLCC/SROCC. However, \cite{xu2016blind} itself lists their performance as 0.7280 SROCC (compare Table VIII, column `All'). These performance improvements are even more substantial, with 0.20/0.21 PLCC/SROCC on LIVE-in-the-wild and 0.23 SROCC on TID2013.

According to \multi{}, different combinations of Inception-modules were studied for \multit{}. The best performance, however, was achieved using all of them. For KonIQ-10k test sets, a PLCC of 0.915 and an SROCC of 0.911 represent the best performance. This is slightly below state-of-the-art of 0.937 PLCC and 0.921 SROCC achieved by KonCept512~\cite{hosu2020koniq}, the model from the KonIQ-10k database's authors. On the artificial datasets, the proposed approach achieved state-of-the-art performance at 0.966 PLCC and 0.965 SROCC for KADID-10k, as well as 0.950 PLCC and 0.951 SROCC on TID2013. In comparison, related works perform notably worse on KADID-10k with 0.876/0.878~\cite{zhu2020metaiqa}, 0.855/0.830~\cite{ding2020image} and 0.938/0.936~\cite{lin2020deepfl}, as well as on TID2013 with 0.910/0.844~\cite{yan2019naturalness}, 0.880/0.879~\cite{zhou2019no} and 0.876/0.858~\cite{lin2020deepfl} at the time of publication. This improvement is, therefore, very notable.

Finally, \deeprnt{} is evaluated on KonIQ-10k, with cross-tests on LIVE and LIVE-in-the-wild, meaning that the model trained on KonIQ-10k was evaluated without re-training on these databases. At the time of publication the KonIQ-10k database had yet to be evaluated by modern IQA approaches, so the authors implemented BosICIP~\cite{bosse2016deep}, CNN~\cite{kang2014convolutional} and \deepbiqt{}. The former two had similar performance at 0.67/0.65 and 0.67/0.63 PLCC/SROCC, respectively, while \deepbiqt{} was reported to be 0.92/0.90 PLCC/SROCC. It is to be noted, that the authors described \deepbiqt{} as using a VGG16 network as a feature extractor, which is incorrect. Their own proposed approach was claimed to perform at 0.95/0.92 PLCC/SROCC, a substantial improvement of 0.03/0.02 over \deepbiqt{} and nearly 0.30 PLCC and SROCC over the other methods. 

 \section*{Data leakage cases}
\label{sec:dataleakageft}

\deepbiqt{} is the first work that described the general approach of fine-tuning, feature extraction, and subsequent training of a regressor \added{for the purpose of IQA}. Later papers that we discuss in the following resolved some of the data leakage cases of the earlier works, while introducing new data leakages. Therefore, we will first discuss data leakage Case I from \deepbiqt{}. Both \svrt{} and \lstmt{} fixed this, but instead introduced two more types of data leakage, which we discuss afterwards. Finally, \multit{} introduced a fourth data leakage case very similar to Case II, which we handle last.

\subsection*{Case I}

In \deepbiq{} \deepbiqt{} was introduced, where a CaffeNet network, pre-trained on ImageNet and Places, is modified by removing the last layers up to `fc7', a fully connected layer of size 4,096, and instead adding a 5-way fully-connected softmax classification layer. Conventionally, for fine-tuning of networks pre-trained on ImageNet images, each channel of an input image is normalized by the subtraction of the mean intensity of the training set. In this implementation, however, each input image $i$ is put through a normalization procedure that is described as "subtracting the mean image that is computed by averaging all the images in the training set on which the CNN was pre-trained". Since the network was pre-trained on both ImageNet and Places, which have different resolutions, it is non-trivial to replicate this normalization procedure. After the input normalization, a random crop of 227x227 resolution is taken and trained on an output $C(i)$ that corresponds to a coarse quantization of the image's original MOS$_i$:

\begin{equation}
C(i) = 
\begin{cases}
\mathrm{Excellent} & \text{if } \mathrm{MOS}_i \in ]80,100],\\
\mathrm{Good} & \text{if } \mathrm{MOS}_i \in ]60,80],\\
\mathrm{Fair} & \text{if } \mathrm{MOS}_i \in ]40,60],\\
\mathrm{Poor} & \text{if } \mathrm{MOS}_i \in ]20,40],\\
\mathrm{Bad} & \text{if } \mathrm{MOS}_i \in [0,20].
\end{cases}
\end{equation}

Additionally, class balancing is used during the backward pass to emphasize weight updates from images belonging to less frequent classes.

The authors describe the fine-tuning process incompletely as being carried out for 5,000 iterations, where a single iteration commonly refers to the forward and backward pass of a single mini batch. A thorough explanation of many vital parameters used in the fine-tuning process is lacking, such as \replaced{details about}{the descriptions of} \added{the} batch size, learning rate, and gradient descent optimizer. \replaced{The authors of \deepbiq{} have communicated that there is no publicly available implementation for \deepbiqt{}, which, coupled with the lack of information on training parameters, }{ This} makes it difficult to reproduce the approach exactly. Furthermore, since the paper does not state utilizing a validation set as an early stopping criterion to avoid over- or underfitting, it is questionable how valid the 5,000 iterations criterion is.

In our reimplementation of this fine-tuning process we attempted to stick as closely to the described procedure as possible, while also employing a validation set, so as to maximize generalization performance. Before the fine-tuning procedure we defined five random training, validation and test splits, which are used throughout the entire implementation. The fine-tuning is carried out on a training set and evaluated on the respective validation set after every epoch. The training process was stopped early, if validation performance did not improve for 25 epochs in a row. For input normalization we chose to employ the conventional way of subtracting the RGB vector [104, 117, 124] for each pixel. With a batch size of 83 and learning rate of 0.001 using stochastic gradient descent the fine-tuning process on LIVE-in-the-wild stopped after an average of 36 epochs ($\pm 7$ standard deviation), with an average validation set classification accuracy of 46.98\% ($\pm 3.09$ standard deviation). 

After fine-tuning, multiple random crops from each image are passed through the network and the activations of the `fc7' layer are extracted, yielding a vector representation for each patch, made up of 4,096 features. \deepbiq{} considers different feature aggregation methods, however we will focus only on the feature pooling approach, which stacks the features of $n$ random patches into a $n \times 4,096$ matrix and averaging along the first dimension, yielding the average image patch feature vector as an image-level feature representation. The paper goes on to describe a 80/20 data split for the training of the SVR regressor. However, there is no clear communication that the splits used for training of the final regressor are the same as in the fine-tuning step. It is important to note, that this is a crucial requirement for a fair evaluation of the method. Therefore, in our reimplementation we split the data prior to fine-tuning and kept the splits fixed throughout the whole learning process. \added{In the case of TID2013 we performed the splits according to the reference image in order to avoid data leakage.} As an input to the model we take $n=30$ random patches, a number found to be an acceptable choice in the original paper. The SVR is then trained on the 80\% of the data that is obtained by joining both the training and validation sets. The final performance is reported in Table \ref{tab:perf_results_deepbiq} in the third row at $0.7886 (\pm0.024$) PLCC and $0.7585 (\pm0.020)$ SROCC on the independent test sets. This is a difference of 0.1140/0.1266 PLCC/SROCC than what was reported in \deepbiq{} (row 1). Evaluating our reimplementation on TID2013 showed an even more pronounced difference in performance, with a final performance on independent test sets of $0.6953 (\pm0.059$) PLCC and $0.6443 (\pm0.054)$ SROCC. This is a drop of 0.2547 PLCC and 0.3057 SROCC.

This performance gap can be explained with a simple case of data leakage. If instead of fine-tuning the network on solely the images of the training set, the network is fine-tuned using the entire dataset, some information about all images in the dataset is baked into the convolutional filters of the entire network. The final 4,096 features extracted for the training of the SVR, therefore contain information about the test set that should not be available if implemented correctly. In this way the purpose of the test set cannot be fulfilled, as it is not an independent sample anymore. Moreover, the longer the fine-tuning process is carried out, the more the original information contained in the networks shifts towards the focus of the fine-tuning task. This means that each additional epoch of fine-tuning exposes the network to information from the training data, which in this case also incorporates the test set data. Crucially, information from the test set are therefore introduced into the network, which subsequently boosts the performance of the final SVR in an illegitimate manner.

By eliminating the validation procedure and fine-tuning on the entire LIVE-in-the-wild dataset for 4900 iterations with otherwise the same settings of above, the final training set classification accuracy of the fine-tuning stage jumps to 100\%, as it is heavily overfitting on the training data. The SVR performance then also increases to $0.9191 (\pm0.018)$ PLCC and $0.9018 (\pm 0.014)$ SROCC, which is very close to the reported 0.9026 and 0.8851 PLCC and SROCC. For TID2013 the final SVR performance increases to $0.9621 (\pm0.003)$ PLCC and $0.9578 (\pm 0.004)$, again close to the reported 0.96/0.96 PLCC/SROCC.

\begin{table}[t!]
  \caption{Performance results of \deepbiqt{} on LIVE-in-the-wild (rows 1-3) and TID2013 (rows 4-6) according to \deepbiq{}, as well as our own reimplementation of the approach as described in the paper. The last column designates whether fine-tuning (column `ft') was performed correctly (green checkmark), or with data leakage Case I (red cross). The numbers in bold font in lines 3 and 6 give the true performance of \deepbiqt{}, much below the claimed 0.90/0.89 PLCC/SROCC for LIVE-in-the-wild and 0.96 PLCC and SROCC for TID2013.}
  \label{tab:perf_results_deepbiq}
  \centering
  \begin{tabular}{rcccc}
    \cmidrule[1pt]{2-5}
     &  src & PLCC \added{($\pm \sigma$)} & SROCC \added{($\pm \sigma$)} & ft \\
    \cmidrule[1pt]{2-5}
     1 & \cite{bianco2018use}  & 0.9026 ($\pm$--.--{}--{}--) & 0.8851 ($\pm$--.--{}--{}--) & \rxmark \\\
     2 &   ours & 0.9191 ($\pm$0.018) & 0.9018 ($\pm$0.014) & \rxmark \\
    \cmidrule{3-5}
     3 &   ours & \textbf{0.7886 ($\pm$0.024)} & \textbf{0.7585 ($\pm$0.020)} & \gcmark \\
    \cmidrule[1pt]{2-5}
     4 &  \cite{bianco2018use}  & 0.9600 ($\pm$--.--{}--{}--) & 0.9600 ($\pm$--.--{}--{}--) & \rxmark \\
     5 &   ours & 0.9621 ($\pm$0.003) & 0.9578 ($\pm$0.004) & \rxmark \\
    \cmidrule{3-5}
     6 &   ours & \textbf{0.6953 ($\pm$0.059)} & \textbf{0.6443 ($\pm$0.054)} & \gcmark \\
    \cmidrule[1pt]{2-5}
  \end{tabular}
\end{table}

It can be observed that the correct implementation performs worse on the artificially degraded IQA database in TID2013, than it does on the authentic IQA database LIVE-in-the-wild, while the reverse is true for the numbers reported in \deepbiq{}. This further suggests that the described case of data leakage is present in \deepbiqt{}. If data leakage occurred for the original implementation, a higher performance at the same number of iterations should be expected for TID2013 over LIVE-in-the-wild, due to the smaller amount of original contents. In each epoch the network is exposed to multiple variations of the same image, shifting the convolutional filters to be more specific to the contents of the few original images. At the same numbers of iteration the network has been overfit on the few contents, allowing higher levels of performance on the artificial dataset. By removing the data leakage the performance drops significantly, because the small amount of original contents is insufficient for the fine-tuning stage to extract generalizable features.

\subsection*{Case II}

The general approaches of both \svrt{} \svr{} and \lstmt{} \lstm{} are very similar to what was described in the previous Section, with a few key differences. \added{Both implementations first fine-tune and then extract features from individual frames of videos, using pre-trained convolutional neural networks. This second case of data leakage occurs in the fine-tuning process, which we will describe in detail first.} 

The author describes the fine-tuning process as follows. A pre-trained Inception-style network is modified, such that the final layer is replaced with a 5-way fully-connected softmax layer, using the Xavier weights initialization. The inputs to the networks are video frames, downscaled and center-cropped. The outputs correspond to the five intervals that contain the video's mean opinion score. Concretely, the class $C(v[i])$ for the $i$th frame of video $v$ as an input to the network is assigned as:
\begin{equation}
C(v[i]) = 
\begin{cases}
\mathrm{VeryGood} & \text{if } 4.2 < \mathrm{MOS}({v}) \leq 5.0,\\
\mathrm{Good} & \text{if } 3.4 < \mathrm{MOS}({v}) \le 4.2,\\
\mathrm{Mediocre} & \text{if } 2.6 < \mathrm{MOS}({v}) \le 3.4,\\
\mathrm{Poor} & \text{if } 1.8 < \mathrm{MOS}({v}) \le 2.6,\\
\mathrm{VeryPoor} & \text{if } 1.0 \le \mathrm{MOS}({v}) \le 1.8.
\end{cases}
\end{equation}

Fine-tuning was performed on batches of 32 input frames using stochastic gradient descent with momentum $\beta = 0.9$ and an initial learning rate $\alpha = 10^{-4}$. The author states that the rate was divided by 10 when the validation loss stopped decreasing during training, although the online code does not do this. 

Both approaches were evaluated on the KoNViD-1k dataset, consisting of 1,200 video sequences with accompanying MOS values. According to both papers, 240 videos were randomly chosen as a test set, put away, and not used during the fine-tuning step. The remaining 960 videos were used for training and validation, splitting the dataset 4:1. As a further subsampling step 20\% of the frames of all 960 videos were randomly selected to constitute the combined training and validation set for the fine-tuning and feature learning. This set of extracted frames was further divided into a training and validation set. Although the paper does not specify what training to validation set ratio was used, it can be assumed that the ratio was 3:1, as an overall 3:1:1 ratio between training, validation, and test sets is common in deep learning.

As a result of the training for the classification task, the author reported in both \svr{} and \lstm{} a classification accuracy on the validation set after fine-tuning that is higher than 95\%. Unfortunately, this high validation accuracy is not achievable when implementing the approach, as described. In fact, to an observer familiar with machine learning, the author's fine-tuning training progress plot (reproduced in Figure~\ref{fig:fine-tune-repro}) raises concern. The quantization of scalar MOS values into five equisized bins introduces unnecessary ambiguity and complexity. Two points stand out:

\begin{figure}[hbt!]
    \centering
    \includegraphics[width=0.99\columnwidth]{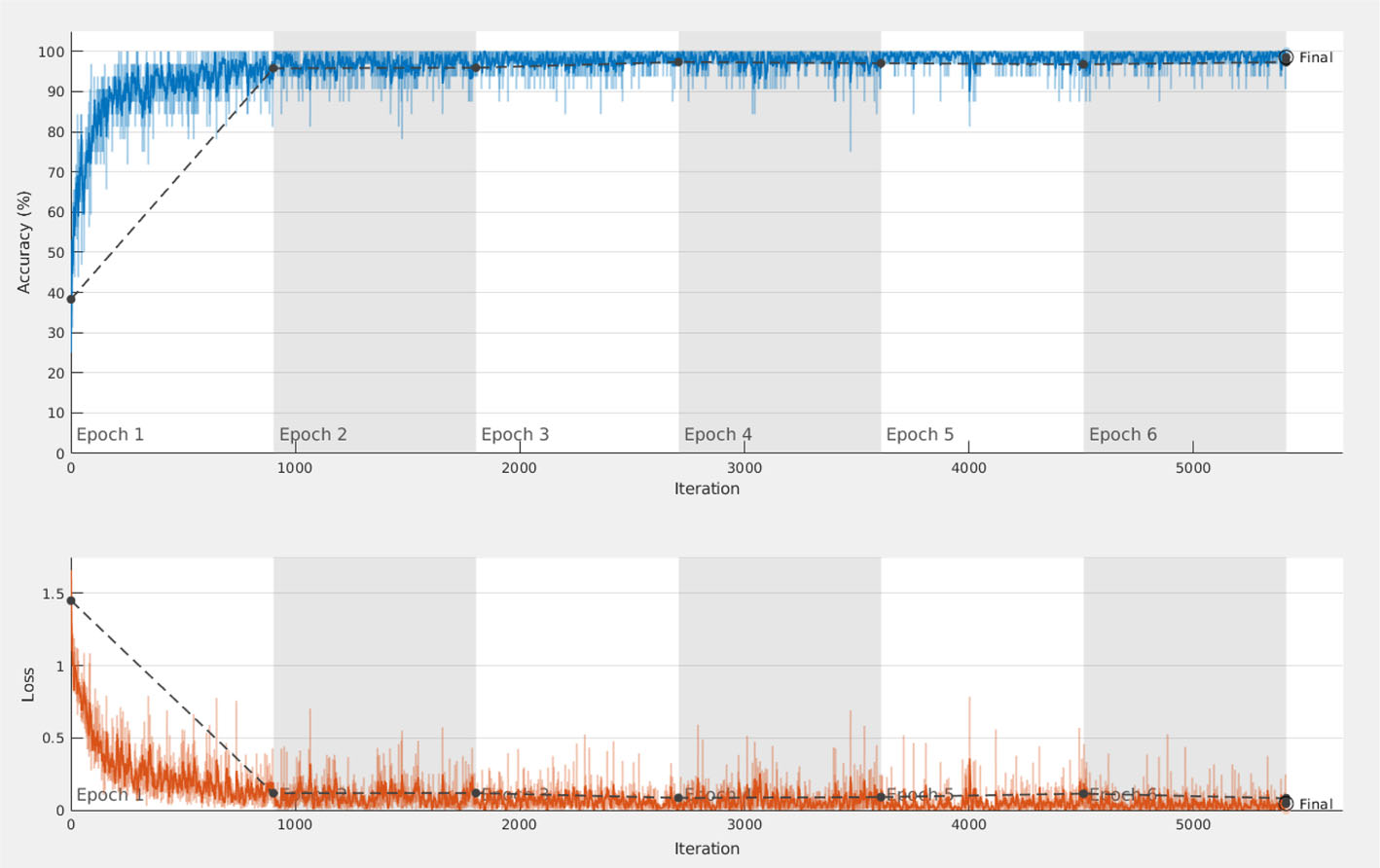}
    \caption{The training progress during fine-tuning as reported in \svr{}. The blue lines show smoothed and per iteration training accuracies in dark and light color variants, respectively. Similarly, the orange lines depict smoothed and per iteration training losses in dark and light color variants, respectively. The dashed dark gray lines linearly connect the validation accuracies and losses indicated by the dark gray circle markers.}
    \label{fig:fine-tune-repro}
\end{figure}

\begin{figure}[t!]
    \centering
    \includegraphics[width=0.99\columnwidth]{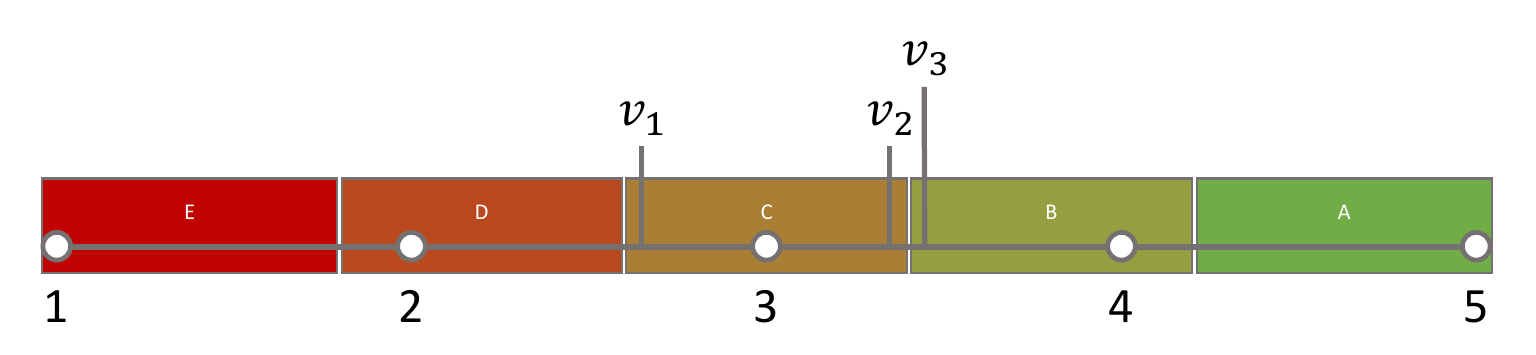}
    \caption{A diagram of the MOS scale (numbers 1 to 5), and the class labels used in \svr{} and \lstm{} for fine-tuning the network represented by the bins E to A. Each bin interval is highlighted by a different color (red to green). The video MOS values are binned according to the five intervals (E to A) to form the target classes used during training. The quantization of MOS values to the five bins reduces labelling precision, which makes training more challenging. For instance, given three videos $v_1$, $v_2$, and $v_3$ at adjacent class boundaries, the difficulty of the classification task becomes apparent. The perceived quality of $v_2$ and $v_3$ is very similar, but they are split into different classes. Conversely, $v_1$ and $v_2$ have a less similar quality than the previous pair, but they are grouped into the same class. }
    \label{fig:class_complexity}
\end{figure}

\begin{enumerate}
    \item The quick increase of both the training and validation accuracy of the training procedure seems unreasonable, given the coarseness of the classes. At class boundaries, the classification task is difficult, as illustrated in Figure~\ref{fig:class_complexity}. Although this increased classification complexity at class boundaries is inherent to all classification tasks, it was unnecessarily and artificially introduced in this case. Based on perceptual information alone, a human would be hard-pressed to perform the classification up to an accuracy of 95\%. It seems very unlikely that the reported classification accuracy on the validation set is achievable in such a difficult scenario.
    \item Complex DCNNs, trained on small datasets, like the one used in this work, eventually overfit if training keeps going on long enough. The validation set is meant as a tool to detect overfitting and, therefore, as a criterion to stop training. Overfitting can be detected by comparing the change in validation set performance. In the onset of overfitting, the gap between training and validation set performance starts widening. The validation set performance improvement stagnates and eventually reverses, while training set performance continues rising. However, in this plot, there is no such \textit{noticeable} stagnation in the validation set accuracy, as the training process runs to the maximum epoch threshold. 
\end{enumerate}

\begin{figure*}[t!]
    \centering
    \includegraphics[width=0.99\columnwidth]{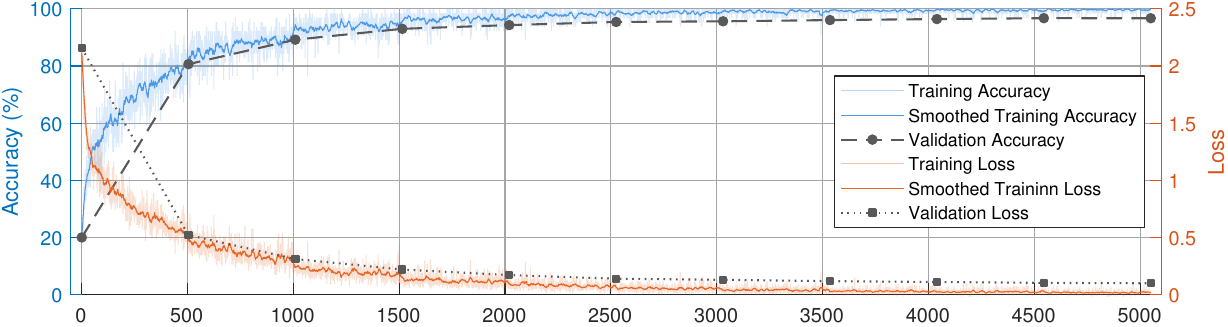} \\
    \includegraphics[width=0.99\columnwidth]{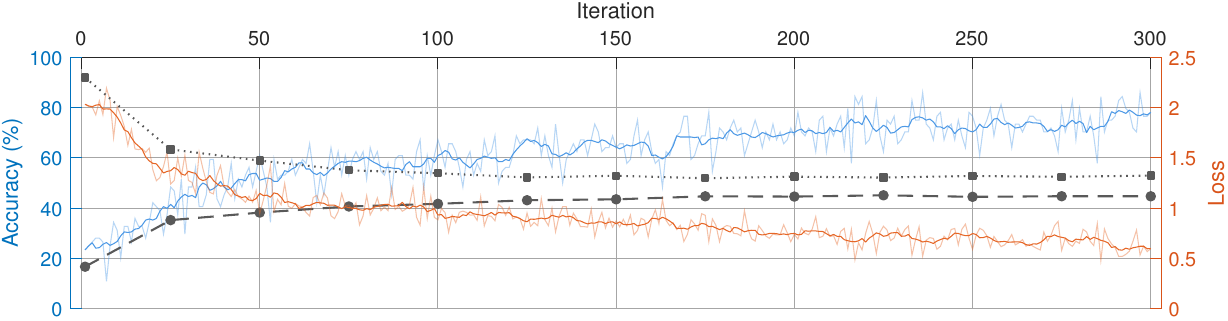}
    \caption{Comparison of reimplementations of the fine-tuning procedure. The top figure depicts the training progress of a fine-tuning procedure with data leakage, while the bottom figure shows the training progress of a fine-tuning procedure without data leakage.}
    \label{fig:fine-tune}
\end{figure*}

Figure~\ref{fig:fine-tune} depicts the training progress of the fine-tuning step. On top is our reimplementation of the author's approach, and its corrected version is shown below. To obtain the plot in the upper part, we had to introduce a particular form of data leakage that can be found in an earlier version of the author's public code\footnote{\urlx{https://github.com/Skythianos/No-Reference-Video-Quality-Assessment-Based-on-the-Temporal-Pooling-of-Deep-Features/tree/621f689eae8319be79af80497db55d97637ea213}} and is therefore likely to have been the cause of this implausible fine-tuning performance. The author was notified of this error in August 2019, as can be seen in the discussion of this problem on the author's code repository issues page\footnote{\added{\urlx{http://web.archive.org/web/20201205103659/https://github.com/Skythianos/No-Reference-Video-Quality-Assessment-Based-on-the-Temporal-Pooling-of-Deep-Features/issues/2}}}.

In the author's original implementation, the first selection of 80\% of the videos was used to fine-tune the feature extraction network. Then 20\% of the frames from these videos were randomly selected and pooled in a data structure. From this data structure, a random selection for training and validation was made. Obviously, this causes frames from the same video to end up in both of the sets, defeating the validation set's purpose. The validation set is meant to be sampled independently from the training set to fairly estimate a model's generalization potential on an independent test set. Since this is not the case, the validation performance does not indicate the performance on an independent test set. The validation and training performances are very similar, as the two sets are nearly identical in content.

Consequently, when the model starts to overfit on the training set, this cannot be detected by the validation procedure used for both models. The fine-tuned models should have lower performance on an arbitrary set of videos independent of the training set, as is the case for the test set. From the earlier versions of the author's code as well as from Figure~\ref{fig:fine-tune-repro}, it can only be concluded that this case of data leakage was present in the particular implementation that was used in \svr{} and \lstm{} to produce the results reported for \svrt{} and \lstmt{}.

In both works discussed here, some training parameters were poorly chosen. Evaluation on the validation set is conventionally performed once per epoch after the entire training data was passed through the network once. If the inputs are independent images, e.g., in an object classification problem, this is a reasonable approach. However, in \svr{} and \lstm{}, the training set consists of 20\% of all frames from each video selected for training. In the case of a 240 frame long video, this amounts to virtually 48 frames being passed through the network before the validation set is being evaluated. Compared to the object classification task on images from above, this is comparable to 48 epochs. As mentioned above, the evaluation of the validation set is used to select the best generalizing model. Infrequent validation can lead to poor model selection. Therefore, we evaluated the validation set more frequently in our reimplementation in order to select the best performing feature extraction model. Validation occurred once every 1600 frames in our training procedure, compared to once every 32,000--33,000 frames in the original implementation. Comparing the two plots in Figure~\ref{fig:fine-tune}, we can see that the training procedure shown in the bottom stops at iteration 300. Here, the validation loss (black dots, dotted line) is not improving anymore, while the training loss keeps decreasing (orange line), which triggers the stopping criterion. However, in the top plot, the first validation set evaluation only occurs after 500 iterations. If we were to employ the same validation frequency, we would likely select a sub-optimal model.

\begin{figure*}[t!]
    \centering
    \includegraphics[width=0.99\columnwidth]{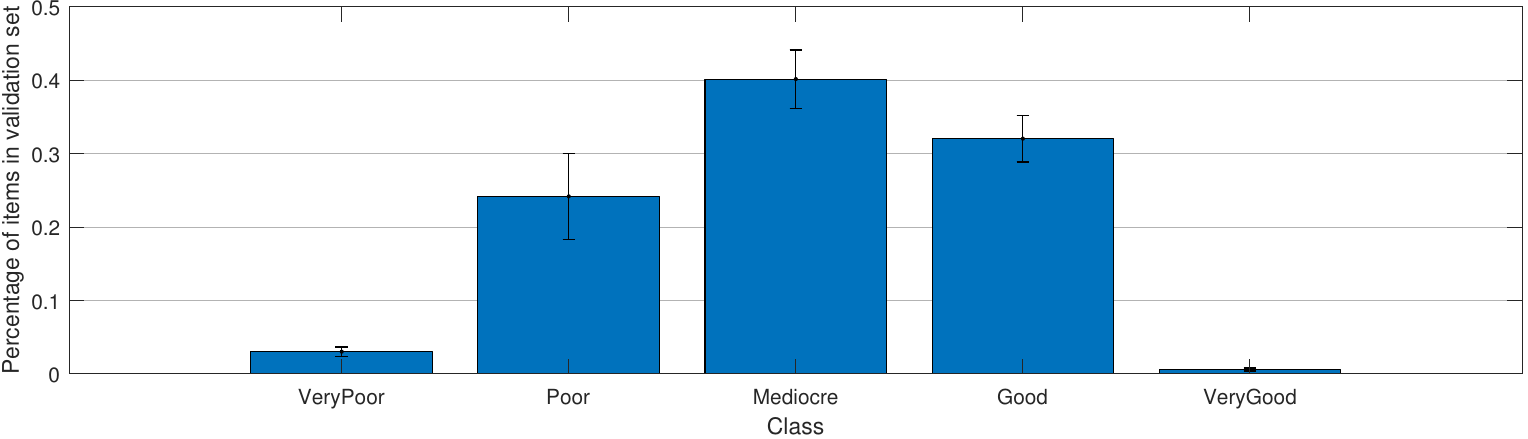}
    \caption{Average distribution of class predictions in percent across the five splits used for the fine-tuning of the feature extraction model. The error bars denote the standard deviation.}
    \label{fig:class_dist}
\end{figure*}

Moreover, the fine-tuning process in itself does not seem to have a significant impact. Figure~\ref{fig:class_dist} shows the distribution of predicted video classes in the test set averaged over five random splits with the error bars representing the standard deviation. The average peak test accuracy for the classification task across five correctly fine-tuned models is $46.52\%$. When simply predicting the dominant class equally for all queries, the average test set accuracy is $41.08\%$. The $5.44\%$ increase in accuracy over the dominant class predictor is a marginal improvement indicating that the classification task may not be appropriate. This could be due to the problems with grouping MOS scores into coarse classes as described earlier or a more general problem of Inception-V3 features not being informative enough about video quality. We investigate the latter in the Discussion section. 

\subsection*{Case III}

After fine-tuning, \added{both \svrt{} and \lstmt{} proceed similarly to \deepbiqt{} with training a final predictor on} the extracted features\added{. In this second phase, an additional case of data leakage occurs, that is somewhat similar to Case I, in that information from the test set used to evaluate the method was already used in the fine-tuning stage. The} extracted features are used as inputs to a model that learns to predict the stimulus's overall visual quality. There is, however, a slight difference between \svrt{} and \lstmt{} in how the features are used for quality predictions, as can be seen in Figure~\ref{fig:flowchart}. Specifically, in the case of \svrt{}, features extracted from individual video frames are aggregated before being input into an SVR architecture. Here, \svr{} considers mean, median, minimum, or maximum aggregation methods to obtain video-level feature representations. For \lstmt{} the stack of feature vectors from individual frames of the same video is used as an input to the LSTM architecture used to predict the video's quality. Both approaches suffer from the same additional case of data leakage, that we will describe in the following.

\subsubsection*{\svrt{}}

\begin{figure*}[t!]
    \centering
    \includegraphics[width=0.99\columnwidth]{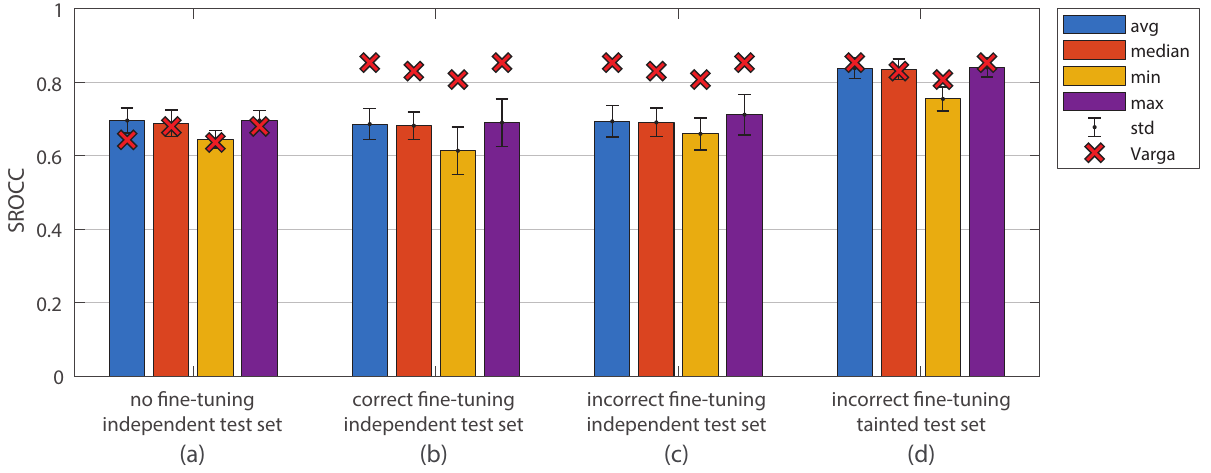}
    \caption{Performance comparison of SVRs trained using different kernel functions from our reimplementation. Chart (a) shows the results when no fine-tuning is used for the feature extraction network. The performance with correctly applied fine-tuning is shown in the chart (b), which is also the \textit{true} performance of the approach. Charts (c) and (d) depict the performance when fine-tuning is performed with data leakage. The bars represent the average performance of five random training, validation, and test splits. Independent test sets are chosen prior to fine-tuning, and for (d) also tainted test sets are chosen at random before SVR training. The red cross markers represent the corresponding numbers reported in \svr{}, as measured from the figures in the paper.}
    \label{fig:results_svr}
\end{figure*}

Figure~\ref{fig:results_svr} (a) shows the average performance of five SVRs trained with a gaussian kernel function without fine-tuning of the feature extraction network as a reproduction of what was reported in \svr{}. The colors indicate four different ways of aggregating frame features into video level feature representations. The approximate results reported in the original publication, as measured from the figures in the original paper, are shown by the red cross markers, and they match those of our reimplementation. In this case, the fine-tuning data leakage described in the previous section has no effect, as no fine-tuning is employed.

Chart (b), on the other hand, shows the performance of SVRs trained on the same splits but with correctly implemented fine-tuning in the first step. More importantly, the SVRs were trained using only the training and validation set videos that were already used in the fine-tuning process. The test set was not made available at the fine-tuning stage nor in the SVR model training.

We see a vast difference in performance between our reimplementation and the performance numbers reported by the author as denoted by the red crosses, which cannot solely be attributed to incorrect fine-tuning. Figure~\ref{fig:results_svr} (c) depicts the average performance values of the five SVRs with incorrect fine-tuning evaluated on the independent test sets with little improvements over the chart (b). This begs the question of what might have happened in the performance evaluation process in \svr{}.

The standard practice when training a machine learning regressor is to utilize k-fold cross-validation. One reports the average performance on models trained on multiple random training, validation, and test splits. This is also just what was done in \svr{}, as the paper explains, ``The different versions of our algorithm'' (different pooling strategies, different SVR kernels) ``were assessed based on KoNViD-1k by fivefold cross-validation with ten replicates in the same manner as the study by \cite{li2016spatiotemporal}.'' Checking the paper \cite{li2016spatiotemporal} confirms that the whole dataset was split into folds, each being used as a test set for the SVR. \added{What has to be ensured in this specific case, though, is that the data used as a test set is not sampled from the set of videos used in the training and validation sets of the fine-tuning stage.} \replaced{Otherwise}{Therefore}, 80\% of the videos in each test set \replaced{would have}{had} already been \replaced{used}{utilized} in the network fine-tuning stage\replaced{, and}{. So} most of the feature vectors from a test set \replaced{would have}{had} been learned in the feature extraction network from their corresponding video MOS values\added{.}\deleted{, and in the end, it was the job of the SVR to predict the same MOS values from these learned features.} This \added{would} constitute\deleted{s} another clear case of data leakage resulting in `tainted' test sets, which \added{could} explain\deleted{s} why our reimplementation \replaced{did}{could} not reach the performance claimed in \svr{}.

Based on the above \replaced{assumptions}{analysis}, we succeeded to reproduce the results published in \svr{} with random splits into training, validation, and \textit{tainted} test sets for the training and testing of the SVR. Figure~\ref{fig:results_svr} (d) corresponds to the average performance of five gaussian kernel function SVRs evaluated on tainted test sets is shown, with the standard deviation denoted by error bars. \added{We believe that the similarity in results obtained by introducing tainted test sets is a strong indication that this type of data leakage did occur in the implementation of \svrt{}.}

\subsubsection*{\lstmt{}}

In \lstm{}, the video frame features were also extracted by passing the frames through the feature extraction network individually. However, since the LSTM architecture used as a predictor in \lstmt{} takes a series of features as input, no aggregation was performed. Instead, the features were concatenated along the temporal axis. They were zero-padded at the end where necessary, as the training process is performed on batches of data with the same length. The training data were sorted according to the number of frames to keep the amount of zero-padding small; the batch size was set to 27. Additionally, the batch selection was set to be non-random, meaning that the same data would appear together and at the same time in the training process. 

\begin{figure}[t!]
    \centering
    \includegraphics[width=0.99\columnwidth]{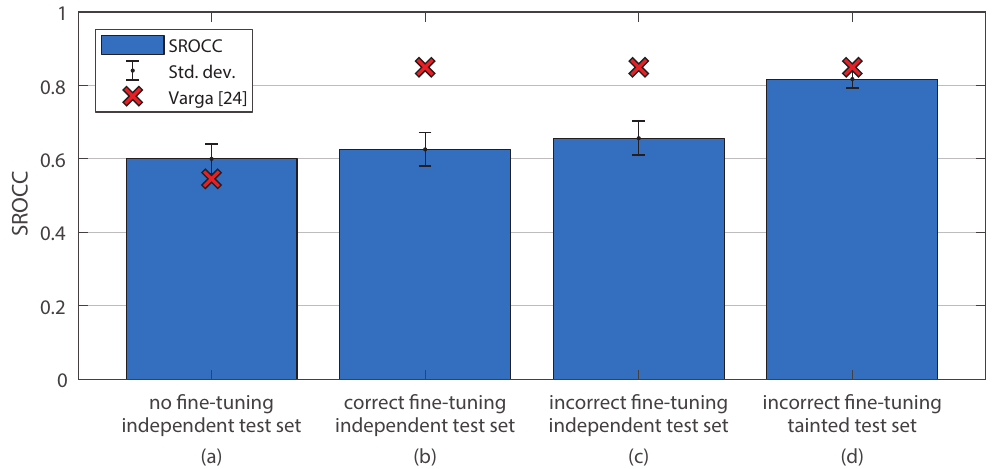}
    \caption{Performance comparison of our reimplementation of the approach described in~\lstm{}. Again, bar (a) depicts the performance when no fine-tuning is used for the feature extraction network. When correctly applying fune-tuning we obtained the performance shown in bar (b), which is also the \textit{true} performance of the approach. Bars (c) and (d), then, indicate the performance when fine-tuning is performed with data leakage. All bars represent the average performance of five random training, validation, and test splits. Independent test sets are chosen before fine-tuning, and for (d) also tainted test sets are chosen at random before SVR training. The red cross markers represent the corresponding numbers reported in \lstm{}, as measured from the figures in the paper.}
    \label{fig:results_lstm}
\end{figure}

We applied the same methodology as for the previous investigation of this approach. Figure~\ref{fig:results_lstm} (a) shows the average performance of five LSTMs trained on features extracted from an Inception-V3 network without any fine-tuning. The red crosses indicate the approximate results reported in \lstm{} as measured from the figures in the original paper. Our reimplementation closely matches the results published in the original publication.

However, as before, when using fine-tuning, the published results substantially deviate from our reimplementation. Figure~\ref{fig:results_lstm} (b) again shows the performance of LSTMs trained on the same five splits as before, such that the training of the LSTM networks and the fine-tuning happens on the same sets, disjoint from the test sets. The test set videos were only used in the final evaluation of the LSTMs. Analogously to the previous discussion, the difference in performance between our implementation and that in \lstm{} cannot be attributed to the incorrect fine-tuning alone, as can be seen by Figure~\ref{fig:results_lstm} (c). Only by using tainted test sets does the performance increase to a level comparable to the published results, as shown in the last chart of Figure~\ref{fig:results_lstm}. The test sets for the LSTM model are tainted because many of the videos contained in them had previously been used for the feature learning in the fine-tuning stage.

\subsubsection*{Summary}

Table~\ref{tab:perf_results_svr} provides a summary of the performance results of various VQA algorithms on KoNViD-1k alongside the results published in both \svr{} and \lstm{} as well as our reimplementation results. The middle section (rows 7 to 10) compares the original approaches without fine-tuning as reported in the original publication and as re-computed by us. These entries correspons to the left plots of Figures~\ref{fig:results_svr} ands~\ref{fig:results_lstm}. As described before, since no fine-tuning was performed, the data set splits have no impact; therefore, test sets can not be tainted with data items that the network had seen before. The performance numbers we obtained are very similar to those reported in both articles. 

\begin{table}
  \caption{Performance results of various VQA algorithms on KoNViD-1k. The data is taken from the references listed in the second column. In the upper half, the first column gives the abbreviated name of the algorithm. The lower half denotes the base architecture used to extract features (column `base') and the model used to predict the overall quality (column `pred'). The last two columns designate whether fine-tuning (column `ft') was performed correctly (green checkmark), or with data leakage (red cross), and whether the test set (column `test') was independent (green checkmark) or tainted (red cross). The two approaches indicated by \ssymbol{1} were published after the referenced publication and are current state-of-the-art. --.--{}-- indicates unreported values. The numbers in bold font in lines 15 and 20 give the true performance of \svrt{} and \lstmt{}, much below the claimed performance.}
  \label{tab:perf_results_svr}
  \centering
  \begin{tabular}{rrccccccc}
    \cmidrule[1pt]{2-9}
        & \multicolumn{2}{c}{VQA algorithm}   & src & PLCC \added{($\pm \sigma$)} & SROCC \added{($\pm \sigma$)} & & &  \\
    \cmidrule{2-6}
    1 & \multicolumn{2}{c}{CORNIA}    & \cite{korhonen2019two} & 0.51 ($\pm$0.02) & 0.51 ($\pm$0.04)   &   &   &  \\
    2 & \multicolumn{2}{c}{V-BLIINDS}  & \cite{korhonen2019two} & 0.58 ($\pm$0.05) & 0.61 ($\pm$0.04)   &   &   &  \\
    3 & \multicolumn{2}{c}{STFC} & \cite{men2018spatiotemporal} & 0.64 ($\pm$--.--{}--) & 0.61 ($\pm$--.--{}--)   &   &   &  \\
    4 & \multicolumn{2}{c}{TLVQM}     & \cite{korhonen2019two}  & 0.77 ($\pm$0.02) & 0.78 ($\pm$0.02)  &   &   &  \\
    5 & \multicolumn{2}{c}{3D-CNN+LSTM\ssymbol{1}}    & \cite{you2020deep}  & 0.81 ($\pm$--.--{}--) & 0.80 ($\pm$--.--{}--)  &   &   &  \\
    6 & \multicolumn{2}{c}{MLSP-VQA-FF\ssymbol{1}}    & \cite{gotz2019no}  & 0.83 ($\pm$0.02) & 0.82 ($\pm$0.02)  &   &   &  \\
    \cmidrule[1pt]{2-9}
    7 & I-V3 & SVR & \svr{}  & 0.72 ($\pm$--.--{}--) & 0.68 ($\pm$--.--{}--)   & max & -  & - \\
    8 & I-V3 & SVR  & ours & 0.69 ($\pm$0.06) & 0.66 ($\pm$0.07)    & max &  - & - \\
    \cmidrule{4-9}
    9 & I-V3 & LSTM & \lstm{}  & 0.51 ($\pm$--.--{}--) & 0.55 ($\pm$--.--{}--) & -  & -  & - \\
    10 & I-V3 & LSTM & ours & 0.62 ($\pm$0.07) & 0.60 ($\pm$0.06)  & -  &  - & - \\
    \cmidrule[1pt]{2-9}
    11 & I-V3 & SVR & \svr{}  & 0.85 ($\pm$--.--{}--) & 0.85 ($\pm$--.--{}--)& avg &\rxmark&\rxmark \\
    12 & I-V3 & SVR  & ours & 0.85 ($\pm$0.01) & 0.85 ($\pm$0.01)& avg &\rxmark&\rxmark \\
    13 & I-V3 & SVR  & ours & 0.72 ($\pm$0.03) & 0.69 ($\pm$0.05)& avg &\gcmark&\rxmark \\
    14 & I-V3 & SVR  & ours & 0.74 ($\pm$0.02) & 0.73 ($\pm$0.02)& avg &\rxmark&\gcmark \\
    15 & I-V3 & SVR  & ours & \textbf{0.70 ($\pm$0.04)} & \textbf{0.67 ($\pm$0.04)} & avg &\gcmark&\gcmark \\
    \cmidrule{4-9}
    16 & I-V3 & LSTM & \lstm{}  & 0.87 ($\pm$--.--{}--) & 0.85 ($\pm$--.--{}--) &-&\rxmark&\rxmark \\
    17 & I-V3 & LSTM & ours & 0.84 ($\pm$0.02) & 0.84 ($\pm$0.01) &-&\rxmark&\rxmark \\
    18 & I-V3 & LSTM & ours & 0.69 ($\pm$0.04) & 0.67 ($\pm$0.04) &-&\gcmark&\rxmark \\
    19 & I-V3 & LSTM & ours & 0.68 ($\pm$0.02) & 0.67 ($\pm$0.02) &-&\rxmark&\gcmark \\
    20 & I-V3 & LSTM & ours & \textbf{0.65 ($\pm$0.05)} & \textbf{0.63 ($\pm$0.05)}&-& \gcmark&\gcmark \\
    \cmidrule{2-9}
        &  base & pred & src  & PLCC \added{($\pm \sigma$)} & SROCC \added{($\pm \sigma$)} & pool & ft & test \\
    \cmidrule[1pt]{2-9}
  \end{tabular}
\end{table}

Next, the bottom half of the table summarizes the results of the approaches including fine-tuning. Here, the last two columns indicate whether fine-tuning was performed correctly (green checkmark) or with data leakage (red cross), and whether the test set was independent (green checkmark) or tainted (red cross), respectively. 

For \svrt{} the reimplemented approach with incorrect fine-tuning and tainted test sets (row 12) closely matches the results reported in \svr{} (row 11). The next two rows 13 and 14 show the individual impact that the two cases of data leakage have. The tainted test sets caused a more significant gap in performance, which was expected, given that this form of data leakage is beneficial to the performance on the test set specifically. Surprisingly, the incorrect fine-tuning appears to improve results over correctly implemented fine-tuning, which deserves additional investigation. 

Row 15 shows the \textit{true} performance of \svrt{}. Both fine-tuning and testing were carried out correctly, with strict training, validation, and test set splitting. The average performance across five random data splits, each fine-tuned using only the training set, model selection performed using the performance on the validation set, and performance reported solely on test set items was 0.70 PLCC and 0.67 SROCC. With this result, the proposed method cannot be considered state-of-the-art, as it performs worse than TLVQM by 0.07 PLCC and 0.11 SROCC, which is a considerable performance gap. Moreover, recent advances in the field \cite{gotz2019no} have pushed performance on KoNViD-1k to above 0.8 PLCC and SROCC, as shown in rows 5 and 6.

Analogously rows 16 to 20 provide the performance numbers on the fine-tuned approaches for \lstmt{} in the same order as before. We reproduce the performance values given in \lstm{}, followed by our reimplementation with both incorrect fine-tuning and tainted test sets in row 17. Next, rows 18 and 19 show the individual impact of tainted test sets (row 18) and fine-tuning with data leakage (row 19) compared to the correct and data leakage-free performance values in row 20. 

In summary, the fine-tuning did increase performance, regardless of whether it was implemented correctly (+0.03/+0.03 PLCC/SROCC) or not (+0.06/+0.07 PLCC/SROCC). Further investigation is required to understand why the presented type of data leakage in the fine-tuning process overall improves performance when it conceptually should not. However, the LSTM-based model performs worse than the SVR, with PLCC and SROCC dropping from 0.70/0.67 (row 15) to just 0.65/0.63, rendering this approach far from state-of-the-art as compared to 3D-CNN+LSTM at 0.81/0.80 or MLSP-VQA-FF at 0.83/0.82 (rows 5 and 6). Moreover, the correct implementation of the approach has a lower performance by 0.22/0.22 from the claimed 0.87/0.85, showing the importance of rigorous evaluation.

\subsection*{Case IV}

In the case of \multit{}, the \added{data leakage, similarly to Case I, occurs in the} fine-tuning process, \added{which} differs slightly from the previous works. Instead of replacing the last layer with a classification head, the Inception-type network is modified to feed into a single-neuron regression layer. Training is done on 20 random\deleted{ly extracted} crops from each of the training set images\replaced{, while}{. T} the label for each crop is the MOS of the source image. The author uses an Adam optimizer on batches of 28 crops with momentum $\beta = 0.9$ and an initial learning rate $\alpha = 10^{-4}$. The learning rate is said to be divided by ten as the validation error plateaus.

Although the author omitted a figure depicting the training progress in \multi{}, his evaluation of some existing works shows large performance discrepancies. Concretely, \multi{} reports HOSA's~\cite{xu2016blind} performance on the entirety of TID2013 at 0.95 SROCC, while the original authors reported it to be much lower at 0.73 SROCC (See~\cite{xu2016blind} Table VIII, column `All'). Also, other classical methods reported in \multi{} showed unusually high correlation coefficients. This difference in performance can be explained by a distinct difference between KonIQ-10k and artificially degraded datasets such as KADID-10k and TID2013, which conventionally contain reference images alongside multiple degraded variants. The KADID-10k database contains 81 pristine images degraded by 25 different distortions with five levels each, resulting in 125 versions of each pristine image. This is similar to the case of adjacent frames from videos, which we discussed above. If two differently degraded variants of the same pristine image are used in both the training and validation set, the image content is largely the same, and the validation set can not accurately indicate generalization performance. Instead, it is a mirror of the performance on the training set. The training, validation, and test sets have to be split according to groups of stimuli when evaluating machine learning models on artificially degraded datasets. All variants of a pristine image have to be grouped into the same set not to run the risk of data leakage.

Unfortunately, the fine-tuning process described in \multi{} only handles the procedure for KonIQ-10k, where randomly splitting images into the training, validation, and test sets is a valid approach. Inspecting code provided\footnote{\label{note1}\urlx{https://github.com/Skythianos/Multi-Pooled-Inception-Features-for-No-Reference-Image-Quality-Assessment}} by the author for \multit{} we found that the code reproduces the published performance numbers for KonIQ-10k. However, when adapting the code to KADID-10k and TID2013 under consideration of the restrictions mentioned earlier, required to avoid data leakage, the resulting PLCC and SROCC values did not match with the published numbers. By randomly splitting images without consideration for the reference image, we achieved the published performance. Therefore, we can conclude that the previously mentioned type of data leakage caused the incorrect performance numbers published in \multi{}.

At the time of publication of \multi{} the state-of-the-art performance of blind IQA on KonIQ-10k was achieved by the KonCept512 model~\cite{hosu2020koniq} with 0.94 PLCC and 0.92 SROCC, closely followed by DeepBIQ~\cite{bianco2018use} with an InceptionResNet-V2 base at 0.91 PLCC and 0.91 SROCC. The proposed MultiGAP-NRIQA model achieved comparable results to DeepBIQ on KonIQ-10k but claimed substantial improvements of 0.05 PLCC and 0.07 SROCC to 0.97 and 0.97 on the artificial dataset KADID-10k. The next best performance on KADID-10k clocks in at 0.94/0.94 PLCC/SROCC~\cite{lin2020deepfl}. Furthermore, they reported their evaluation of related works on TID2013, showing a competitive performance of 0.95 PLCC and 0.95 SROCC on TID2013.

The author provided code for evaluating \multit{} on KonIQ-10k on his personal GitHub repository\added{$^{\ref{note1}}$}. The resulting performance values match what was published. However, the stark differences between their re-evaluation of existing works on KADID-10k and TID2013 and the initially published performance values of these existing works indicate an error. As mentioned previously, a simple random splitting of all images of artificially degraded image quality datasets into training, validation, and test sets causes data leakage. A model will see semantically highly similar images in the training and validation set, and, more importantly, the test set will not be independent of the training or validation sets. The resulting test set performance will not represent the performance on an unbiased and independent set of data.

We adapted the author's original code to KADID-10k, comparing the case where the splitting into the training, validation, and test sets was done randomly to a split based on the grouping of degraded versions of reference images. The procedure was evaluated both with and without fine-tuning. Since the author's code uses a single split throughout the entire process, there is an increased potential for data leakage. This can come from the fine-tuning procedure, similar to what was described in the previous section, as well as from training of the SVR in the end. Table~\ref{tab:perf_results_multi} is a summary of our findings. The first row is a reproduction of the performance values of \multit{} without fine-tuning. Row 2 is our adaptation of the author's original code to KADID-10k, without considering the semantic similarities of degraded versions of the same reference image, indicated by the red cross in the `split' column. The performance values of our five reproducible splits are practically identical to what was published. However, when splitting the dataset according to reference images, performance drops significantly, as shown in row 3. Here, all versions of the same reference image were grouped into the same set in the training process, eliminating data leakage.

\begin{table}[t!]
  \caption{Performance results of \multit{} on KADID-10k alongside our reimplementation, both with and without data leakage. The last column designates whether the splits were sampled correctly considering content (green checkmarks) or randomly, thereby giving rise to tainted test sets (red crosses). The numbers in bold font in lines 3 and 6 give the true performance of the method in \multi{} both with or without fine-tuning, much below 0.97/0.94 PLCC and 0.97/0.94 SROCC, respectively, as claimed. In line 7 we report the results of an implementation of an end-to-end regression network that combines the feature selection and the quality score prediction.}
  \label{tab:perf_results_multi}
  \centering
  \begin{tabular}{cccccc}
    \cmidrule[1pt]{2-6}
         & src & PLCC \added{($\pm \sigma$)} & SROCC \added{($\pm \sigma$)} & ft & split \\
    \cmidrule[1pt]{2-6}
    1 & \multi{}  & 0.94 ($\pm$--.--{}--{}--) & 0.94 ($\pm$--.--{}--{}--)  & - &  \rxmark \\
    2 & ours & 0.94 ($\pm$0.002) & 0.94 ($\pm$0.002) & - &\rxmark \\
    3 & ours & \textbf{0.61 ($\pm$0.074)} & \textbf{0.61 ($\pm$0.081)} & - & \gcmark \\
    \cmidrule{4-6}
    4 & \multi{}  & 0.97 ($\pm$--.--{}--{}--) & 0.97 ($\pm$--.--{}--{}--) & \gcmark&\rxmark \\
    5 & ours & 0.97 ($\pm$0.002) & 0.97 ($\pm$0.002)   & \gcmark & \rxmark \\
    6 & ours & \textbf{0.81 ($\pm$0.047)} & \textbf{0.81 ($\pm$0.043)} & \gcmark & \gcmark \\
    7 & ours & 0.80 ($\pm$ 0.026) & 0.81 ($\pm$ 0.027) & \gcmark & \gcmark \\
    \cmidrule[1pt]{2-6}
  \end{tabular}
\end{table}

Next, the bottom part of the table represents the performance of the approach with fine-tuning. Row 4 again reproduces the performance values of \multi{}, while row 5 shows our adaptation to KADID-10k with random splitting. The correct way of splitting the data again impacts performance, as shown in row 6, although not as strong as without fine-tuning. Fine-tuning the Inception-V3 network with the input images' quality scores using a regression head improves the usability of the features for the SVR training. Nonetheless, with 0.81 PLCC and 0.81 SROCC, the final performance is still far from state-of-the-art at 0.94/0.94 PLCC/SROCC~\cite{lin2020deepfl}. 

\subsection*{The Case of \deeprnt{}}
Our investigation of \deeprnt{} differs from the previous publications in two major ways. The first is that there is a partial overlap in authorship between the article introducing \deeprnt{} \deeprn{} and this paper. In fact, this is the root cause of our initial investigation, as our own reimplementation of \deeprnt{} proved to be fruitless in obtaining comparable performance
. Secondly, during personal communication with the first author of \deeprn{} we learned that \deeprnt{} included a data leakage similar to Case I, i.e. it was fine-tuned on the entirety of KonIQ-10k, the core dataset this method was evaluated on. 

We attempted to reproduce the results shown in \deeprn{} by introducing this Case I data leakage. The method involved, similarly to the previously discussed approaches, fine-tuning a deep architecture in the first stage, extracting features and training another model on these features. Following the author's description we fine-tuned the model on the entire KonIQ-10k dataset. The second stage was carried out with independent training, validation, and test sets. The code used is an extension of the one provided in the KonIQ-10k GitHub repository\footnote{https://github.com/subpic/koniq}. In addition to introducing the data leakage, the following changes were made:
\begin{itemize}
    \item The initial learning rate is set to 0.01, as in the DeepRN paper. The code in the KonIQ-10k GitHub repository had used a lower starting value (0.0001), which achieves a better performance when the correct training procedure is applied, but which reduces performance in the case of the data-leakage.
    \item The number of epochs, before the learning rate is divided by 10, is set to 50 instead of the 100 epochs used in the KonIQ-10k repository. This helps when saving a limited number of intermediate models (every 10 epochs), extracting features and testing the performance of the second-stage trained model.
\end{itemize}

\begin{figure}[h!]
    \centering
    \includegraphics[width=0.80\columnwidth]{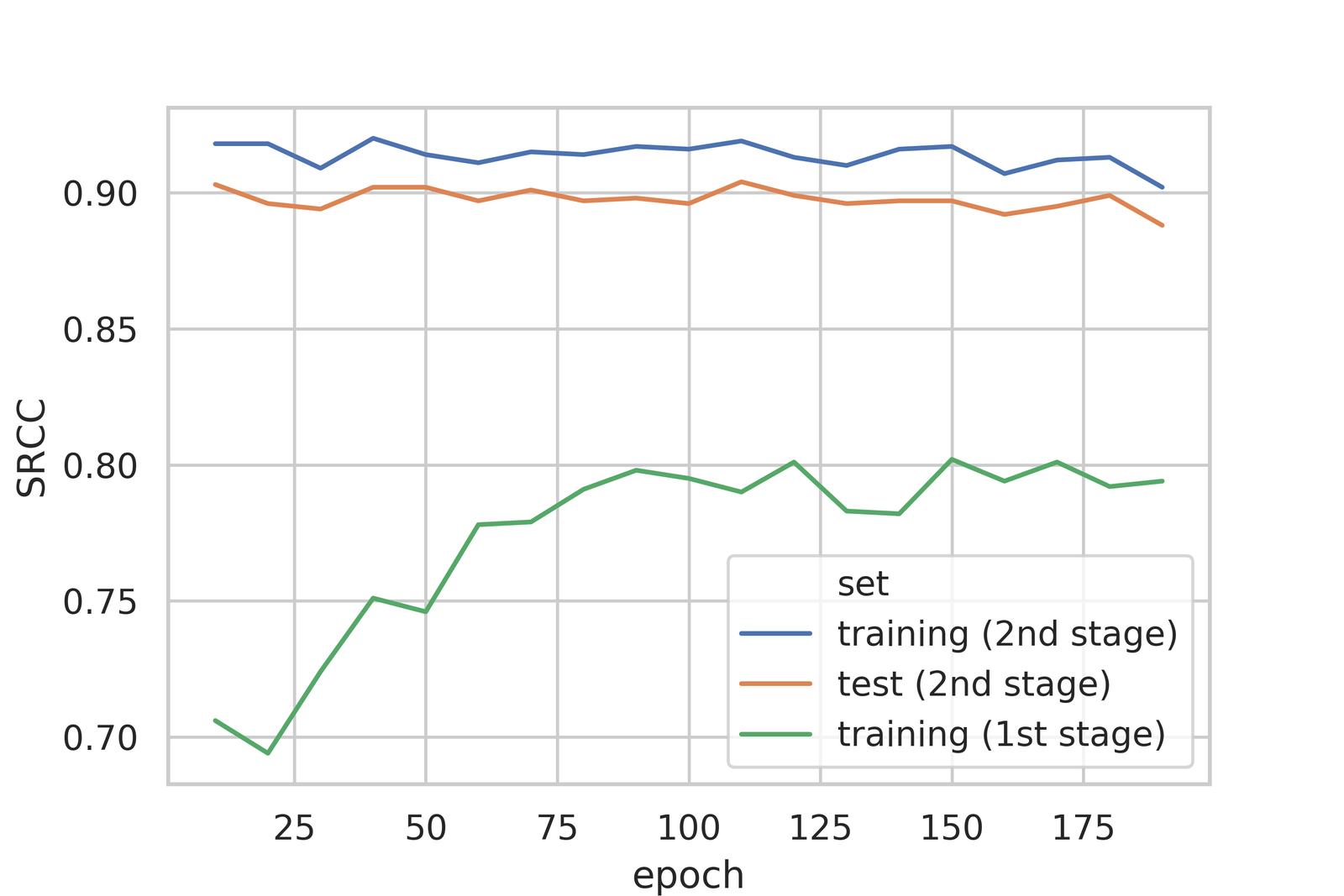}
    \caption{Performance (SROCC) of the DeepRN \deeprn{} model under data-leakage Case I. The network is trained in two stages. In the first stage the network is trained on the entire KonIQ-10k database for the number of epochs shown on the x-axis. Intermediate models are saved every ten epochs, and the second stage training is performed on each of them by correctly using the official training/validation and test sets. The first stage models are trained using the cross-entropy loss, outputting five class-likelihoods, one for each score range. In order to easily compare the reported performances of the two stages, we calculate the corresponding scalar scores from the class-likelihoods. The likelihoods form a distribution over the respective integer scores. Thus, we compute the SRCC between the means of these distributions and the corresponding ground-truth MOS. Neither the test, nor the training performance in the second stage matches the results reported in the DeepRN paper.}
    \label{fig:deeprn-performance-dataleak}
\end{figure}

In Fig.\ \ref{fig:deeprn-performance-dataleak} we show the performance of the models in both stages. For the first stage of the training, we see no improvement in performance beyond epoch~120. This means that at that point we have trained for a sufficient number of epochs. For the second stage, the performance generally decreases at the end of the training, as the first stage model is starting to overfit. The effect is stronger in the final epoch, where the test performance (0.888 SROCC) with the data-leakage is similar to the one reported in the KonIQ-10k paper \cite{hosu2020koniq}, 0.867 SROCC, without a data-leakage.

Our test set results show that introducing the data-leakage does not reproduce the published results in the DeepRN paper \deeprn{} (0.92 SROCC on the test set). The training set performance generally sets an upper bound on the test set performance. Our results show that in the absolute best case, which cannot generally be achieved in practice, the test set performance could go as high as 0.92 SRCC (the best training set performance).  


Therefore, we cannot give a satisfying explanation for what else may have caused the illegitimate performance values published in \deeprn{}. However, we are certain that the results are wrong. Together with the other authors of \deeprn{}, we have notified the IEEE to this extent about these issues, urging a formal expression of concern to be attached to the publication.

\section*{Discussion}
\label{sec:discussion}

Among the different data leakage cases, Case I is the closest to the traditional interpretation, where information about the test set is made available at training time. The other cases are mostly caused by challenges related to the structure or nature of the data. In this particular form, Case II can only occur in the VQA domain, as singular images are made up of only a single frame. However, a similar problem would arise, if in the case of \deepbiqt{} the authors had extracted the random crops first and only then performed the data splits. This would have created the opportunity to omit splitting according to the source image, and it would have been comparable to Case II. Case IV is related to, but not the same as Case II. Here, the different artificial degradations applied to the source image can be understood to be similar to different frames from the same video. However, the VQA domain also has artificially degraded video datasets. Therefore both Case II and Case IV could co-occur in the same implementation and therefore have to be separated. Finally, Case III is most similar to Case I. The similarity between the splits used in the evaluation of the final predictor is reduced from 100\% of fine-tuning training data to 80\% of fine-tuning training or validation data. In fact, when searching for implementations of \deepbiqt{}, one of the top results is a third-party GitHub repository\footnote{\urlx{https://github.com/zhl2007/pytorch-image-quality-param-ctrl}} that reimplemented \deepbiqt{} for the purpose of estimating optimal parameter settings for webcam image quality. Analyzing this code reveals a type of data leakage very similar to Case IV. Here, the data splits for fine-tuning, and the SVR training are separated, with the fine-tuning being done on a random 80/20 split and the SVR training being performed on a deterministic 80/20 split by using a fixed random state. For this implementation, no test set was held out, as the goal was a robust model. Therefore, the resulting two splits are non-identical, and data from the training set of the fine-tuning process will be included in the validation set of the SVR training stage. This shows how easily data leakage is introduced when implementing multi-stage machine learning methods.

Beyond the problems described in the previous investigations, there are some concerns with the approaches in general. First, a support vector machine (SVM) is not an inherently scalable machine learning approach. Specifically, two characteristics of SVMs are problematic for scale:

\begin{itemize}
    \item The memory storage requirements for the kernel matrix of SVMs scale quadratically with the number of items and
    \item Training times of traditional SVM algorithms scale superlinearly with the number of items.~\cite{si2017memory}
\end{itemize} 

There are approaches to circumvent these problems, but for large-scale feature spaces with many data instances, SVMs commonly train slower and perform worse than simpler approaches. The dimension of the feature space of the inputs used here for VQA is close to a problematic size for SVMs to handle. Moreover, SVR is sensitive to model hyperparameters~\cite{tsirikoglou2017hyperparameters,ito2003optimizing}. Careful hyperparameter optimization is commonly performed to ensure the robustness and reproducibility of the results.

Furthermore, it is not clear why \deepbiqt{}, \svrt{} and \multit{} were split into two separate stages. Instead of having fine-tuned on coarse MOS classes in the case of \deepbiqt{} and \svrt{}, one could have replaced the head of the Inception-style network with a regression head like in \multit{}. Additionally, any use of a regression head would eliminate the need for the SVR stage, as the resulting model is trained to predict the quality of the input feature vector. The feature extraction steps in the discussed publications all utilize some for of kernel activation pooling. However, any type of pooling effectively removes information that could be leveraged in a regression setting. If there are performance gains in training an SVR on the extracted features, the end-to-end training approach should at least be a baseline to compare to. 

We have evaluated the end-to-end training procedure on the five random splits used throughout this article for completeness. Following the approach of \svr{}, we took an Inception-V3 network, removed the layers beyond the last pooling layer, and attached three fully connected layers of sizes 1024, 512, and 32, each followed by a rectified linear unit layer that clips negative values to zero and a dropout layer with 0.25 dropout. The fully connected layers of the new head were trained at a ten times increased rate compared to the rest of the network. This improves the training as the head layers' weights are randomly initialized, while the rest of the network is pre-trained. Lastly, we added a fully connected layer of size 1. We trained this network with stochastic gradient descent with momentum and a learning rate of $\alpha=10^{-4}$ and otherwise default training settings, except for a custom learning rate scheduler, that multiplies the learning rate by 0.75 after every epoch. The network was trained for 10 epochs on 20\% of the frames of videos, to retain comparability to the results in Table~\ref{tab:perf_results_svr}.

For testing, the network's prediction was computed for every frame of the test videos. A video-level score was computed as the average frame-level prediction, resulting in $0.66 (\pm 0.02)$ PLCC and $0.65 (\pm 0.03)$ SROCC. This shows that the two-staged approach proposed in \svr{} was successful in improving video quality prediction over this na\"ive approach.

Additionally, we evaluated the end-to-end baseline approach for \multi{} by evaluating the fine-tuned Inception-ResNet-v2 model as a quality score predictor on the five random splits used throughout this article. Although the performance values on KADID-10k, as shown in row 7 of Table~\ref{tab:perf_results_multi} are not improved, careful hyperparameter tuning could potentially boost performance.

\section*{Conclusion}
\label{sec:conclusion}
In this paper, we have tried to reproduce the presumably outstanding performances of machine learning approaches published in \svr{} and \lstm{} for no-reference video quality assessment, as well as in \deepbiq{}, \multi{}, and \deeprn{} for no-reference image quality assessment. The originally reported performance numbers for reference datasets were well above the state-of-the-art at the time of publication. However, our implementation of the proposed methods, based mostly on the author's code, showed the real performance is far below the claims in the articles.

We have shown four individual cases of data leakage that have likely occurred in the original implementations. By introducing these data leakage errors in our reimplementation, we consistently reproduced the incorrect performance values, as they were published in four of the five discussed publications. Moreover, we brought strong arguments for the claim that the original implementations were affected by these errors, both by inspecting the code published by the author and by careful examination of the experimental setup description.

The subtlety with which these types of data leakage found their way into the machine learning systems of the discussed publications stands as a testament to the working code requirement that reproduces the performance values in a publication. A simple mistake, such as incorrectly splitting data, can bring about drastically different results and potentially hinder scientific progress. 

Beyond our findings, we brought up issues with the way these methods were described in writing. Specifically, fundamental information required for an accurate reimplementation was either omitted or reported insufficiently. 

We urge venues and reviewers to ensure that authors are required to provide accurate information on all hyperparameters used for the training of deep learning models. At the very least, the explicit data splits the models were trained on should be shared, so that any kind of data leakage can be identified clearly and easily. This would avoid incorrect performance values to be published in the future. We hope that our paper will help raise awareness of the danger of insufficient scrutiny concerning possible data-leakage situations.



%
%
%


\end{document}